\documentclass[]{JHEP3} 

\usepackage{epsfig,multicol,bbm}

\newcommand\etal {{\it et al.}}

\preprint{arXiv:1006:xxxx}

\title{Dimensionally Constrained Symbolic Regression}

\author{Suyong Choi\\
Department of Physics\\
Korea University, Seoul 136-713\\
Republic of Korea\\
E-mail: \email{suyong@korea.ac.kr}}

\received{\today} 		
\accepted{\today}		

\preprint{}

\abstract{ We describe dimensionally constrained symbolic regression
which has been developed for 
mass measurement in certain classes of events in high-energy physics (HEP).
With symbolic regression, we can derive equations that are well known in HEP. 
However, in problems with large number of variables, we find that by constraining the 
terms allowed in the symbolic regression, convergence behavior is improved.
Dimensionally constrained symbolic
regression (DCSR) finds solutions with much better fitness than is normally
possible with symbolic regression. In some cases, novel solutions are found.
}

\begin{document}

\section{Introduction}
Extraction of a physical parameter from data is an area where application
of symbolic regression can be useful, especially if the relationship
between the parameter of interest and measured variables are contrived and non-linear \cite{symbregr}.
In the problem of mass measurement of a particle in high-energy physics, 
the relationship is determined exactly if all the decayed particles are
detected and measured by the detector. The problem becomes non-trivial
if some of the particles escape detection.

In $W$ boson mass measurement with $W\rightarrow\ell\nu$ at hadron colliders,
the neutrino ($\nu$) escapes detection, but its transverse components of the momentum are measured
indirectly. In this case, transverse mass ($M_{T}$) is known to be the most sensitive
variable to the $W$ boson mass $M_W$ \cite{mt}.

In searches of interest at LHC and
the Tevatron, such as the Higgs particle and supersymmetry phenomena, 
there are two or more particles which escape the detector without leaving
any signal. It makes mass measurement of these particles challenging.
In some of cases, there many solutions are known, but it is not clear in what sense
they are optimal for the mass measurement 
\cite{dittmarmass, choihwwmass, lesterhwwmass}.
By applying symbolic regression to these problems, we may be 
able to derive optimal equations sensitive to mass and gain new insight.

\section {Dimensionally Constrained Symbolic Regression and Its Implementation}

Symbolic regression(SR) is a method which can be
used to derive relationship between input variables and the desired output values.
It is an application of genetic programming technique. In SR, an individual of 
a population is symbolic form of an equation. By applying genetic operations,
  the population evolves and gradual improvement of the overall population 
  as well its most fit individual occur. Selection is done
at the individual level at each generation based on its fitness. 

We need to define the representation, functions/operators to be used, terminals 
that will be used to form an individual in SR.
These, in addition to genetic operations, fitness function, and numerous parameters,
  define an SR.  Later, we will describe the dimensionally constrained symbolic regression (DCSR)
used for mass reconstruction problems in HEP.
To have the flexibility DCSR as an option, we implemented a general SR in C++ language
based on ROOT C++ libraries \cite{ROOT}.

\subsection{Representation and Genetic Operations in SR}
\begin{figure}
\begin{centering}
\includegraphics[width=0.3\textwidth]{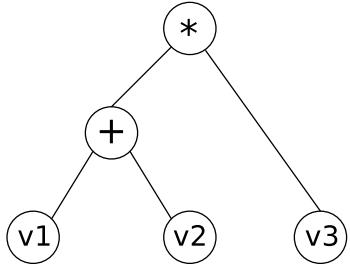}
\par\end{centering}

\caption{\label{fig:A-binary-tree}A binary tree of expression $(v1+v2)*v3$.}

\end{figure}

Internally, mathematical expression is represented as a binary
tree (Fig. \ref{fig:A-binary-tree}). Operators or functions  appear at tree nodes and arguments to them appear as branches.
Binary tree representation allows for efficient manipulation and evaluation of mathematical expressions.

The functions and operators allowed are as follows:
\begin{itemize}
\item Basic arithmetic operators: $+$, $-$, $\times$, $\div$
\item Negation
\item Power
\item Transcendental functions: $\sqrt{ }$, $\log$, $\sin$, $\cos$, $\exp$
\item Unit step function
\end{itemize}

For a function or operator that accepts only one argument, the second leaf may be populated, but 
is not evaluated. Although the second argument has no 
meaning with respect to the function or operator, it may nonetheless affect evolution. 
With cross-over operations and mutations, this second leaf may be picked up for genetic operations.
Our implementation is flexible enough such that we can add more functions or operators, if needed.

When creating an arbitrary expression, the functions or operators are picked
randomly according to certain probabilities. The probabilities should be
provided as an input to SR. To a certain degree, they will depend on the problem to be solved.
We can control the relative rates these functions or operators
are chosen through the input parameters of SR.

\subsection{Terminals}
The input variables and constant become terminals of expression trees. 
In our implementation of SR, 
the relative rate for variables and constants can be controlled. 
However, we set the probability to choose among the variables the same.
For the DCSR, which we will describe latter section, we only allow dimensionless arbitrary constants.

\subsection{Building Expressions from Functions/Operators and Terminals}
A random expression can be build by first choosing a function or an operator randomly with the
predefined probability. This becomes the head node of a tree. Then the leaf nodes are built recursively
by creating a new expression. An expression can be any of the following:
\begin{itemize}
\item Terminal - constant or variable
\item Unary function - $f(x)$, where $x$ is another expression
\item Binary function - $g(x,y)$, where $x$ and $y$ are expressions
\end{itemize}
Since the expression can go on forever, probability to choose a terminal must not be 0, and/or 
when the maximum depth of a tree is reached, one of the terminals is chosen.

\subsection{Parameters of SR}
Below, we list the complete start up parameters
\begin{itemize}
\item Number of variables ($n_{var}$) to use in terminals.
\item Number of generations ($n_{gen}$) in evolution of population.
\item Population size ($n_{pop}$) - At each generation, the population size is kept fixed.
\item Tournament size ($n_{tourn}$) - A parent is chosen through a tournament from a randomly chosen subset of size $n_{tourn}$ from the population.
\item Flag whether to try double tournament $b_{doublet}$ - If $b_{doublet}=true$, each parent is chosen in
separate tournaments. Otherwise, the best two in a single tournament are chosen as parents.
\item Maximum depth of tree ($m_{depth}$).
\item Copy reproduction ($n_{copy}$) - best $n_{copy}$ individuals of a population are simply copied to the next generation. 
These individuals may still participate in sexual reproduction.
\item Fraction of new children produced through sexual reproduction in population ($P_{xover}$) - Number of children in each generation is $n_{pop}\times P_{xover}$.
\item Mutation probability ($P_{mut}$) - Mutation probability per node 
\item Drop probability ($P_{drop}$) - Probability per node to drop it in a mutation
\item Probability to pick a constant ($P_{const}$)
\item Probability to pick $i$ th function or operator ($P_{op,i}$)
\end{itemize}

For the examples in the next section, some of the parameters of interest are: $P_{xover}=0.5$,
$P_{mut}=0~0.1$, $n_{pop}=500~2500$, $n_{tourn}=0~0.2\cdot n_{pop}$.
The basic arithmetic operators $(+,-,\times,\div)$
have probabilities between 5\% to 25\% each. 
The other functions and operators take up the remaining probability equally.
These probabilities vary for each run. 

\subsection{Initialization, training, fitness evaluation and evolution of SR}
In the initialization stage, random expressions are created to populate the first generation.
The user can choose to create an initial population
with the expression filled to the maximum tree depth or not. This 
depends on the problem being considered. 
For the purpose of mass reconstruction, we find that having the initial population filled
to the maximum depth of the tree is not necessarily beneficial, since
most of the initial population has poor fitness and get discarded after
a few generations.

No distinction is made between the training and testing sample. One half of
the input sample is randomly chosen for the training sample
and the other half is the testing sample. This choice is made for every
generation. Fitness of each individual is evaluated on the testing sample according
to the criteria for a problem.

Evolution of population is done through two main avenues. First is
the sexual reproduction, where parents are not chosen randomly, but through
a tournament. In a tournament, random subpool of $n_{tourn}$ individuals are chosen
randomly and the best two individuals are chosen as parents in a single-tournament method.
In a double tournament method, the parents are chosen from two tournaments by selecting
the best fit individual in each tournament.

Once the parents are selected, then the subnodes are selected from two parents and 
swapped in place. We accept only one child from a set of parents, this choice is random. 
It is well-documented
in literature that selecting a random subexpression for swapping
leads to trivial or minimal modifications as
nodes that have terminals are chosen more often. Therefore, 
      we choose internal nodes 90\% of the times for swapping.

Asexual reproduction is done by having mutations occurring in random nodes in
the expression tree. Selected node is either replaced with another expression 
or it is dropped. If it is to be replaced, an expression tree randomly
generated using the parameters of the symbolic regression algorithm. An individual 
which successfully had a child can still participate in an asexual reproduction.
We finally, let the ``elite'' ($n_{copy}$) to go to the next generation without modifications.

\begin{figure}
\begin{centering}
\includegraphics[clip,width=0.5\textwidth]{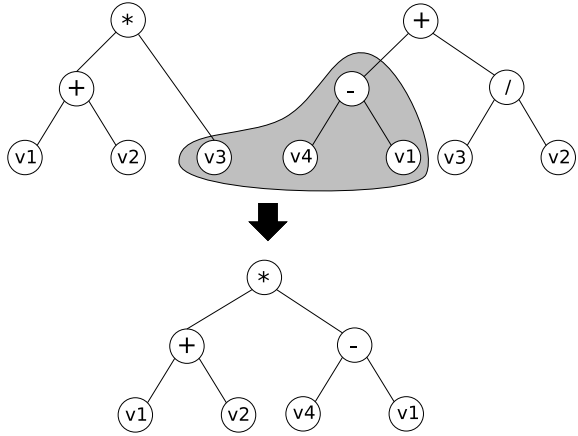}
\par\end{centering}

\caption{\label{fig:xoverop}Creation of a child through cross-over operation.
The cross over of genes occuring between $v3$ and $v4-v1$ from the
two parents yields a new child $(v1+v2)*(v4-v1).$ }

\end{figure}

\subsection{Evaluation of Fitness}
Fitness function is problem specific. Evaluation of fitness is done on a set
of test sample by traversing the expression tree. This is computationally the most time
consuming part of SR.
Expressions which do not yield machine-real
numbers are assigned a number corresponding to a very poor fitness, marking them
likely for removal. An individual with a more compact expression is favorable and
a small penalty proportional
to the number of nodes is imposed on each individual. 

\subsection{Dimensionally Constrained Symbolic Regression}
Symbolic regression must yield a dimensionally sensible result if it is to
be interpretable. In this application, we would like to construct equations
related to mass of a particle. In natural units ($c=1$ and $\hbar=1$),
the mass has the same physical dimensions as momentum and energy. 
We will assume that the target dimension is $p^n$, i.e., 
momentum raised to integer power, and that variables have dimension 1 or 0.
         
We see that the arguments to operators or 
functions must satisfy certain constraints. If $d(a)$ stands for the
physical dimension of expression $a$:
\begin{itemize}
\item $a\pm b$: $d(a\pm b)=d(a)=d(b)$.
\item $a\times b$: $d(a\times b)=d(a)+d(b)$.
\item $a\div b$: $d(a\div b)=d(a)-d(b)$.
\item $\sqrt{a}$: $\frac{d(a)}{2}=d(\sqrt{a})$
\item $x^a$: $d(x^a)=d(x)*a$ and $d(a)=0$ 
\item Transcendental function $f(a)$: $d(f(a))=0$ and $a=0$.
\end{itemize}
Note that the argument to a transcendtal function must be dimensionless.

Using these rules, we can impose that an expression
built by symbolic regression should have the correct physical
dimensions. We would need to reformulate the transformation rules for
evolution compared to the regular SR. 

An $n$ (an integer) dimensional term can be built recursively by applying
one of the following rules:
\begin{itemize}
\item $n-1$-dim term $\times$ $1$-dim term
\item $n$-dim term $\pm$ $n$-dim term
\item $n$-dim term $\times$ $0$-dim term
\item square root of an $2n$-dim term
\end{itemize}
A 0-dimensional term can be created by dividing two 1-dimensional terms or
applying a transcendental function to argument of 0-dimension.
And -1-dimensional term can be created by taking an invervse of 1-dimensional term.
One can see that any integer dimensional term can be created using these rules.

In DCSR, rules for cross-over and mutation need to be modified accordingly.
Exchange or replacement of expressions can occur only among those with the same physical dimensions,
otherwise it is not permitted.

\section{Application of Symbolic Regression to Mass Measurement Problem}

In this section, we will apply symbolic regression to a few problems of mass determination in high-energy physics.
In the first two cases, we apply it to the problem where the answer is well-known. 
And in the third case, we try to find a symbolic expression for the Higgs mass
that can be applied to a case where Higgs boson decays to $W^+W^-$ and each $W$
boson decays leptonically.

\subsection{Invariant mass}
As a first example, we choose a trivial example, one for which we already know the answer.
The variables given are the components of 4-vector, $(p_x,p_y,p_z,E)$, and the target value we would like
to obtain is $m^2=E^2-p_x^2-p_y^2-p_z^2$. A 1000 toy data sample was generated satisfying
the relationship.

\begin{figure}
\begin{centering}
\includegraphics[width=0.4\textwidth]{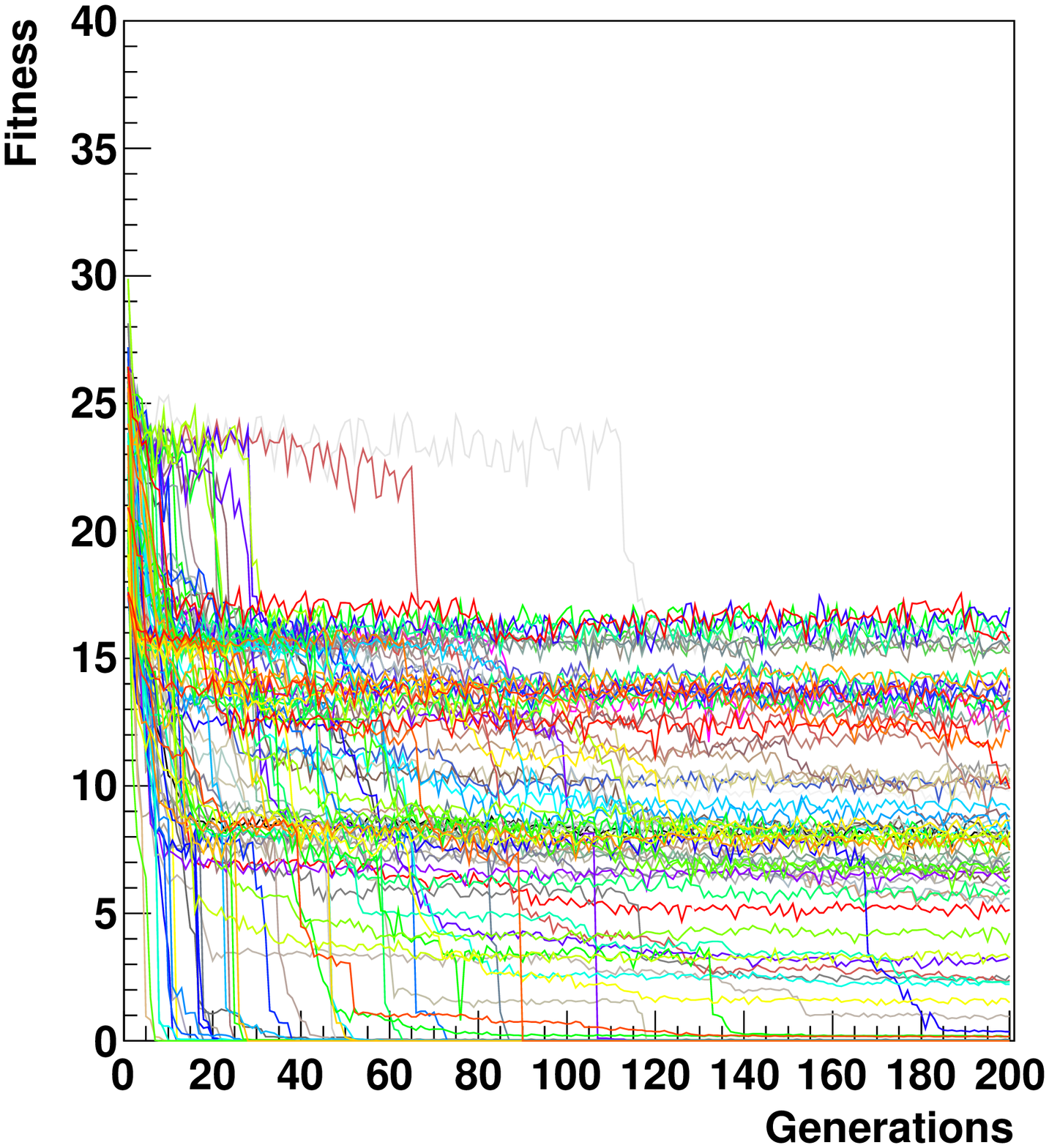}\includegraphics[width=0.4\textwidth]{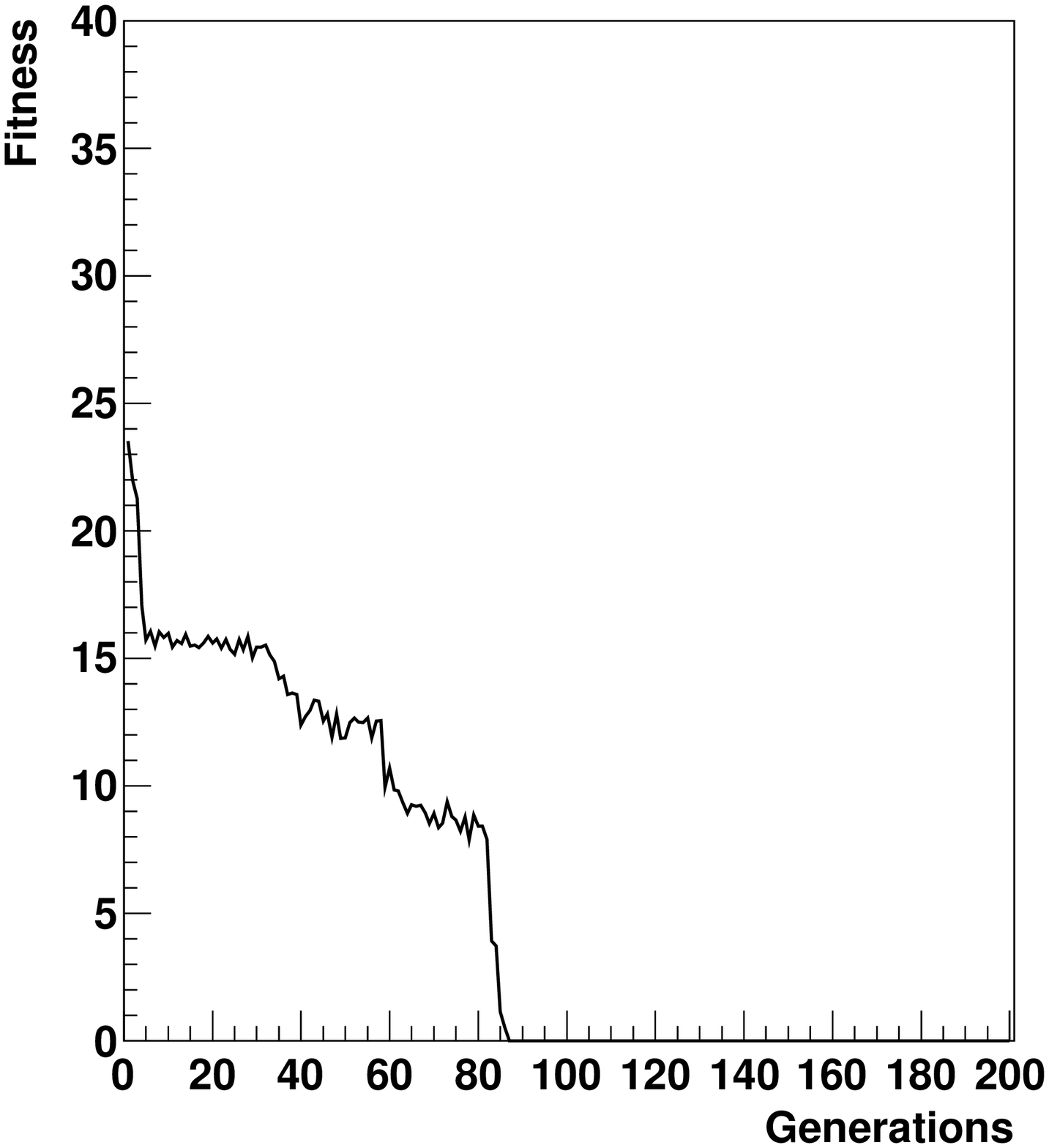}
\par\end{centering}

\caption{\label{fig:Evolution-of-minimum}Evolution of minimum absolute error
of symbolic regression trained with invariant mass data. Out of 100
runs, 26\% yield the correct formula $m^{2}=E^{2}-p_{x}^{2}-p_{y}^{2}-p_{z}^{2}$.
Evolution of fitness of 36th run is shown by itself.}
\end{figure}

Symbolic regressions with different values of parameters were tried
on this data. 
Fitness function is $\frac{1}{n}\sum_{n}|m-m_{true}|$,
where $n$ is the size of data. 
Figure \ref{fig:Evolution-of-minimum}
shows evolution of fitnesses of the best fit individuals in each run.
In 26\% of the runs, the correct formula for invariant mass is found.
The right plot of Fig. \ref{fig:Evolution-of-minimum} shows the evolution
of the best fit individual of run 36. At the start of the run, the
best fit expression is $\sqrt{|E^{2}\cos p_{z}-E|}$, which is meaningless
physically. By generation 86, the equation $m^{2}=E^{2}-p_{x}^{2}-p_{y}^{2}-p_{z}^{2}$
is found. When DCSR is used on the same sample, the chances for success 
increases to 68\%.

\subsection{Events with Missing Information - Transverse Mass}

A more realistic and interesting application of symbolic regression
is to a problem where information is lost. A classic example
would be where a massive particle decays into a charged lepton and
a neutrino in a hadronic collision. Transverse mass $M_{T}=\sqrt{p_{T\ell}\not\not\!\! E_{T}-\vec{p}_{T\ell}\cdot\not\!\!\vec{p}_{T}}$ is typically used to measure the mass of the $W$ boson. To test whether we can
extract this mass relationship, a phase-space generator is used to
create the momenta of lepton and neutrino from a hypothetical massive
particle with an exponentially falling initial transverse momentum distribution for particle
masses between 0 GeV and 100 GeV. 

\begin{figure}
\begin{centering}
\includegraphics[width=0.4\textwidth]{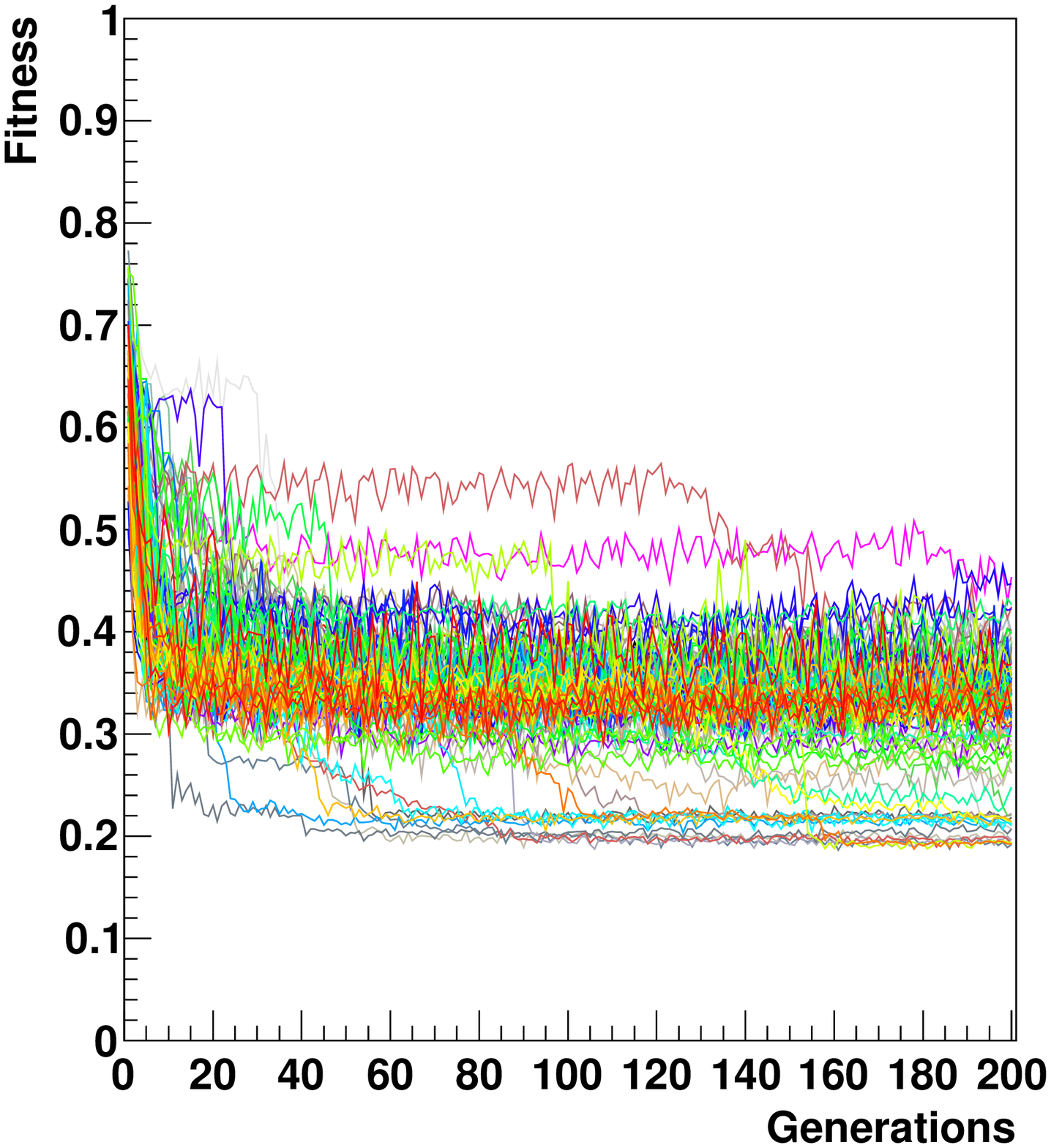}\includegraphics[width=0.4\textwidth]{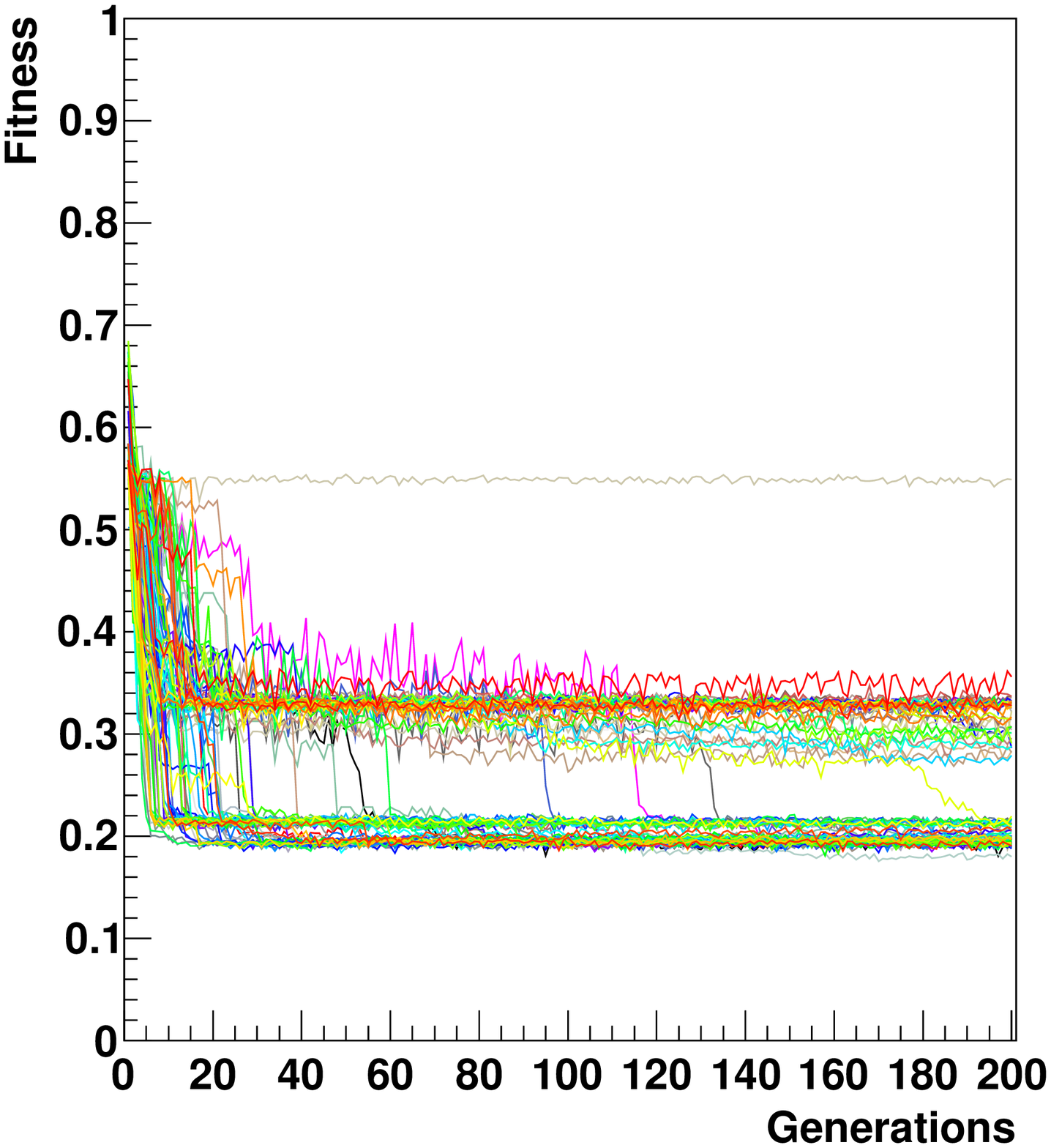}
\par\end{centering}

\caption{\label{fig:mtrun}Evolution of minimum of fitness in 100 runs without
DCSR (left) and with DCSR (right) during training using lepton and
neutrino momenta variable set as inputs $(p_{T},p_{x},p_{y},p_{z},\not\!\! E,\not\!\! E_{Tx},\not\!\! E_{Ty})$.
With DCSR, the fraction of runs that find the equation for transverse
mass increases dramatically.}

\end{figure}

The fitness function used was $\frac{1}{n}\sum^{n}|m_{i,est}-m_{i,true}|/m_{i,true}$,
where $m_{i,est}$ is the value returned by $i^{th}$ individual.
Some trees reach a global minimum as shown in Fig \ref{fig:mtrun}.
For those that reach a global minimum, the equation is $m_{est}^{2}=2.38(p_{T\ell}\not\not\!\! E_{T}-\vec{p}_{T\ell}\cdot\not\!\!\vec{p}_{T})$,
which is the same as the definition of $M_{T}$ except for the overall constant
factor. 

Depending on the parameters of the genetic algorithm, it may take
longer to reach this minimum. Some may fail to reach this minimum
For example, if the minimization criteria were $\sqrt{\frac{1}{n}\sum^{n}(m_{i,est}-m_{i,true})^{2}}$,
almost all trials of symbolic regression fail to reach the global
minimum. The populations get quickly trapped into a local minima.
This is the result of using a minimizing criteria that heavily penalizes
the outliers. It is a known feature of genetic algorithm that having
individuals with poor fitness in the gene pool is important
in population to reach the global minimum. Without DCSR, 7\% 
of the trials find the solutions, while with DCSR the success rate
increases to 40\%.  We find that constraining the terms allowed increases the speed of 
convergence and the chances for a successful convergence.

\subsection{Mass Sensitive Variable in $H\rightarrow WW^{*}\rightarrow\ell^{+}\nu\ell^{-}\bar{\nu}$ }

Higgs mass determination in $H\rightarrow WW*\rightarrow\ell^{+}\nu\ell^{-}\bar{\nu}$
in hadron colliders is an inportant problem. In this channel,
two lepton momenta $\vec{p}_{\ell1},\vec{p}_{\ell2}$
and the vector sum of the two neutrino transverse momenta $\not\!\!\vec{E}_{T}=\not\!\!\vec{E}_{T\nu1}+\not\!\!\vec{E}_{T\nu2}$
are measured in experiments. 
Since there are only two equations related to neutrino momenta, 
the system is under-constrained. If we knew both $W$
bosons were real, we would still need two extra equations to constrain
the system. Therefore one cannot solve for the neutrino momenta
exactly even in principle. 

Existing studies relied on analysis of kinematics to find expressions
that behave linearly to the Higgs boson mass \cite{lesterhwwmass,choihwwmass}.
In this study, we apply symbolic regression to the problem and find an expression
that not only shows linear behavior, but also whose widths of the
mass distribution are narrow. 

Symbolic regression is applied to a data generated with 
PYTHIA $pp\rightarrow H\rightarrow WW^{*}\rightarrow\ell^{+}\nu\ell^{-}\bar{\nu}$
at $\sqrt{s}=14$ TeV with $M_{H}$ varying from 120 GeV to 200 GeV
\cite{PYTHIA}. Detector simulation is not performed on this data in order to find the
ideal expression. Momentum
components and energy of the two charged leptons ($p_{1T},p_{1x},p_{1y},p_{1z},E_{1},p_{2T},p_{2x},p_{2y},p_{2z},E_{2}$)
and missing $E_{T}$ information ($\not\!\!E_{T},\not\!\! E_{x},\not\!\! E_{y}$)
are used as input variables for the symbolic regression. The fitness
function used is the average of fractional absolute difference: $\frac{1}{N}\sum_{i}|M_{rec,i}-M_{H,i}|/M_{H,i}$. 

Without DCSR, the symbolic regression is not able to yield meaningful results. 
This seems to be due to the greater number of variables used.
The number of terms of dimension 2 with only multiplication
allowed is 78, which makes the possible function space to explore very large. 

If fractional root mean-squared (RMS) ($\frac{1}{N}[\sum_{i}(M_{rec,i}-M_{H,i})^2/M_{H,i}^2]^{1/2}$) were used as the fitness function, the symbolic regression 
would get trapped into local minima even with DCSR since outliers pay a heavy penalty. 

Figure \ref{fig:masspred} shows evolution of fitness of best-fit individuals in 100 runs as
a function of the number of generations. DCSR is able to converge
on meaningful results and yields the best estimate for the $M_{H}^{2}$
as \[
S_{mass}^{2}=2p_{1T}^{2}+2p_{2T}^{2}+3\left(p_{1T}p_{2T}+\not\!\! E_{T}(p_{1T}+p_{2T})-\not\!\!\vec{E}_{T}\cdot(\vec{p}_{1T}+\vec{p}_{2T})-2\vec{p}_{1T}\cdot\vec{p}_{2T}\right).\]
Symmetry of the two leptons in the system is recognized by the symbolic
regression automatically, even though symmetry condition was not imposed. 
More detail on the Higgs mass measurement using this result is discussed in Ref. \cite{mysymbr}.

\FIGURE{
\epsfig{file=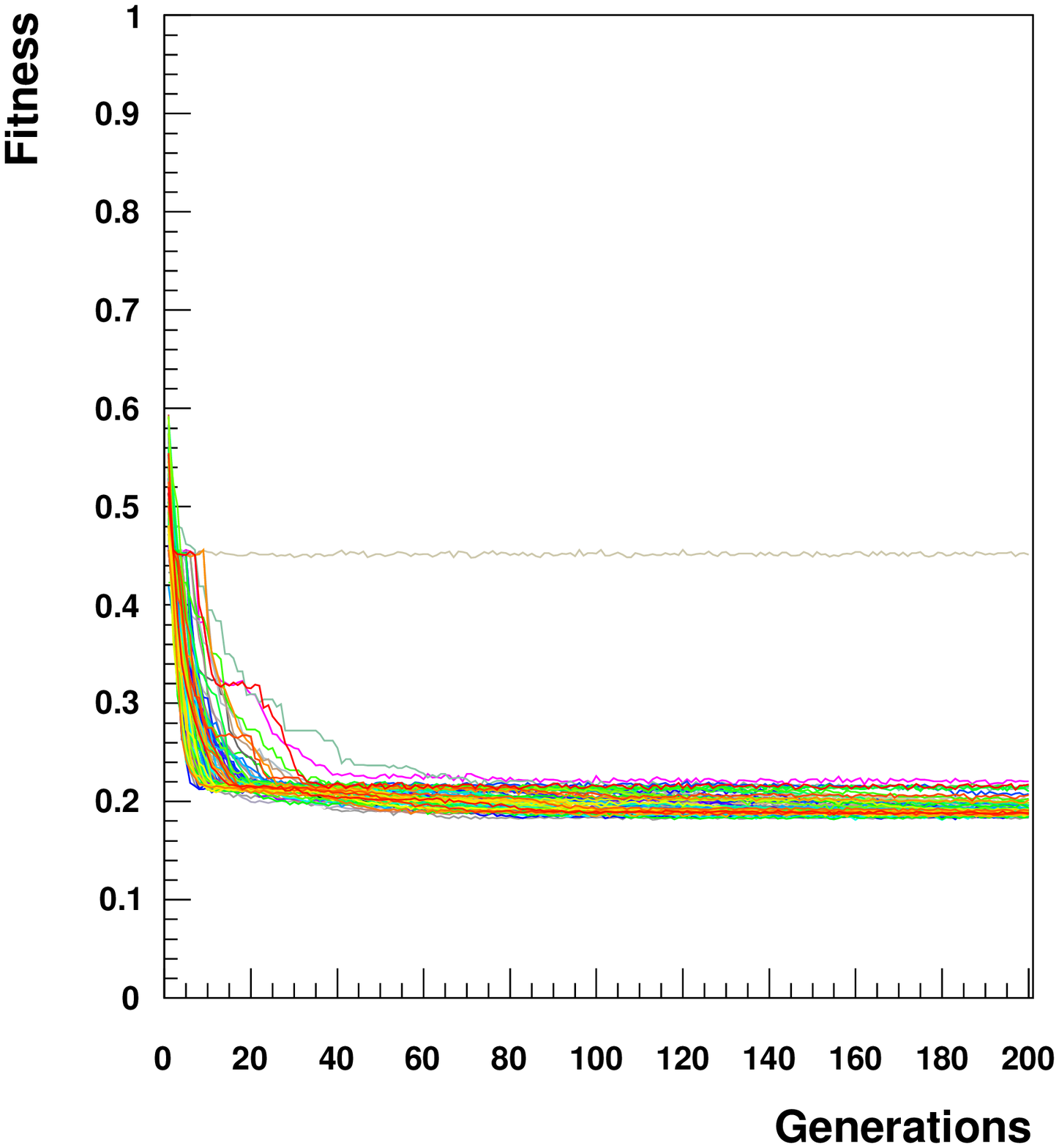,width=0.4\textwidth}
\epsfig{file=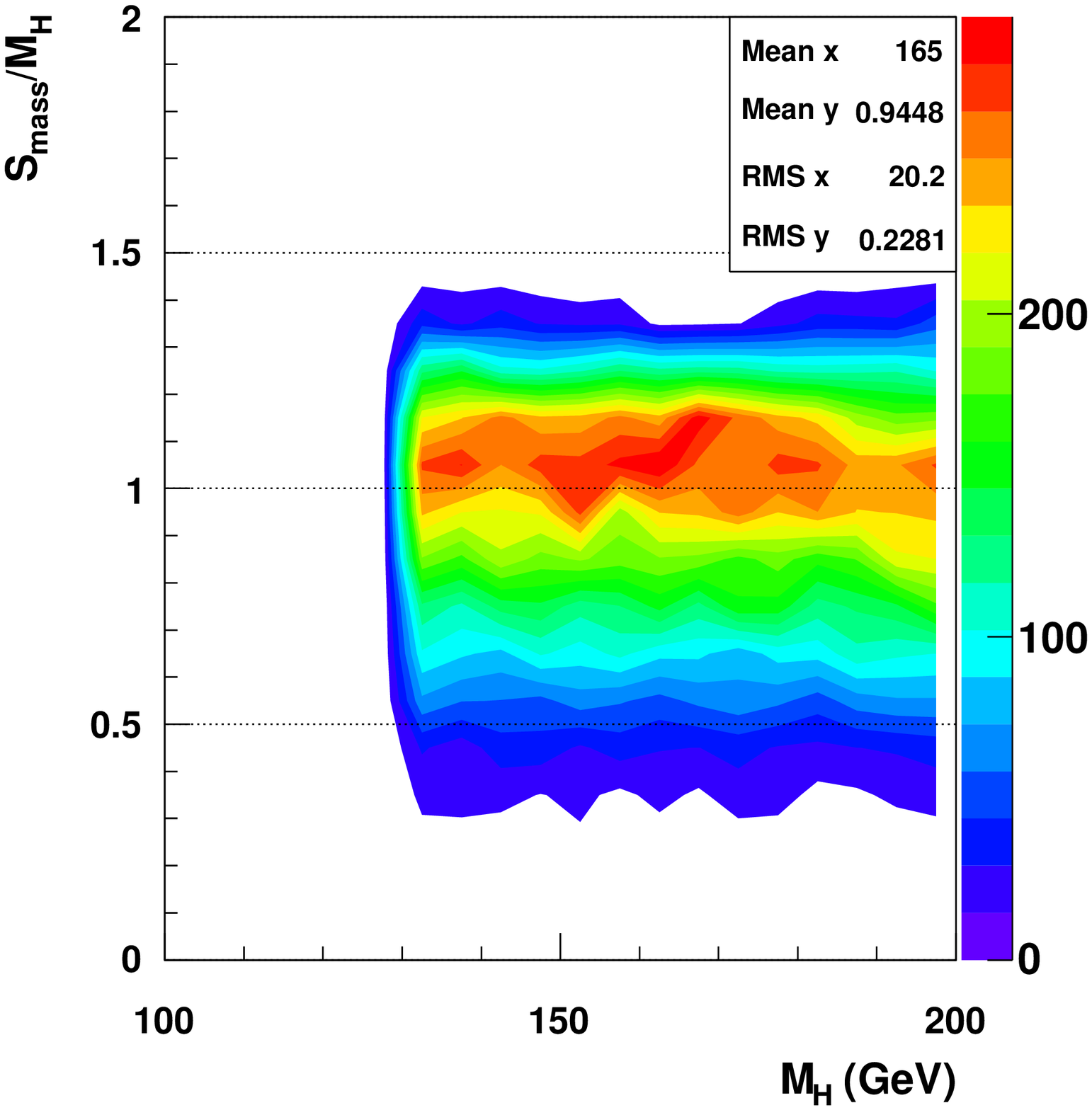,width=0.4\textwidth}
\caption{\label{fig:masspred} Left: Evolution of best fit individuals function
in 100 runs. Right: Distribution of ratio of predicted mass to true
mass, $m_{pred}/m_{H}$, versus the true Higgs mass $m_{H}$. The
sample was generated using PYTHIA.}
}




\subsection{Computational performance}
With our implementation, on a 24 thread X5650 Xeon machine, it takes approximately 1 day to
complete the 100 runs for the Higgs mass example. Most of the time is spent
in evaluation of the fitness function. We noticed memory fragmentation
occurring and evaluation tended to become slower as time went on. A Mathematica implementation,
which was simple to write, did not suffer from memory problems. However, it was inherently slow.
Using the parallel processing capabability of Mathematica 7, even when we used all 24 threads, 
one run took about a day to complete.

With the advent of higher multiplicity core CPUs in the future, it may become
computationally favorable to use this tool. One avenue that needs to be explored is to consider
the backgrounds when optimizing. At the moment, it would be very expensive
to evaluate fitness including backgrounds since a simple fitness function
would no longer suffice.

\section{Conclusion}
We have defined and implemented dimensionally constrained symbolic regression for use in
high-energy physics problems. We find that it reproduces some of the well-known
results, such as invariant mass and transverse mass. We used it
to construct a function that is linear and sensitive to the
Higgs boson mass $M_H$. that has not been previously known.
We expect that this method would be useful in cases with more than one undetected particles, where it
is not trivial what variable could give the best performance.


\begin{thebibliography}{999}
\bibitem{mt}G. Altarelli \etal \npb{246}{1984}{12}.
\bibitem{symbregr}J. R. Koza, 1992. {\it Genetic Programming: On the Programming
of Computers by Means of Natural Selection}, Cambridge, MA: MIT Press.
\bibitem{lesterhwwmass}Alan J. Barr, Ben Gripaios, Christopher Gorham
Lester, \jhep{0907}{2009}{072}.
\bibitem{dittmarmass} M. Dittmar, H. K. Dreiner, \prd{55}{1997}{167}.
\bibitem{choihwwmass}Kiwoon Choi, Suyong Choi, Jae Sik Lee, Chan
Beom Park, \prd{80}{2009}{073010}.
\bibitem{naturalevolution}Holland, John H. 1975. {\it Adaptation in Natural
and Artificial Systems}, University of Michigan Press. 
\bibitem{ROOT}\href{http://root.cern.ch}{http://root.cern.ch}
\bibitem{PYTHIA}Torbj\"orn Sj\"ostrand, Stephen Mrenna and Peter Skands,
\jhep{0605}{2006}{026}.
\bibitem{mysymbr} S. Choi, \arXivid{1006.4998}.
\bibitem{mathematica} Wolfram Research, Inc., Mathematica, Version 7.0, Champaign, IL (2010).
\end{thebibliography}
\end{document}